\documentclass[10pt,twocolumn,letterpaper]{article}

\usepackage[pagenumbers]{cvpr}   

\usepackage{graphicx}
\usepackage{amsmath}
\usepackage{amssymb}
\usepackage{booktabs}

\definecolor{cvprblue}{rgb}{0.21,0.49,0.74}
\usepackage[pagebackref,breaklinks,colorlinks,allcolors=cvprblue]{hyperref}

\usepackage[capitalize]{cleveref}
\crefname{section}{Sec.}{Secs.}
\Crefname{section}{Section}{Sections}
\Crefname{table}{Table}{Tables}
\crefname{table}{Tab.}{Tabs.}

\newcommand{\myparagraph}[1]{\medskip\noindent\textbf{#1}}
\newcommand{\stdev}[1]{{\tiny{±#1}}}

\title{Multi-entity Video Transformers for Fine-Grained\\Video Representation Learning}

\newcommand*{\affaddr}[1]{#1}
\newcommand*{\affmark}[1][*]{\textsuperscript{#1}}

\author{\begin{tabular}{ccc}
  \makebox[.22\linewidth]{Matthew Walmer\affmark[1]\thanks{Work performed during an internship with Meta.}} & 
  \makebox[.22\linewidth]{Rose Kanjirathinkal\affmark[2]} &
  \makebox[.22\linewidth]{Kai Sheng Tai\affmark[2]} \\
  \makebox[.22\linewidth]{Keyur Muzumdar\affmark[2]} &
  \makebox[.22\linewidth]{Taipeng Tian\affmark[2]} &
  \makebox[.22\linewidth]{Abhinav Shrivastava\affmark[1]}
\end{tabular}\\
\affaddr{\affmark[1]University of Maryland, College Park} \qquad
\affaddr{\affmark[2]Meta}\\
}

\begin{document}
\maketitle

\begin{abstract}
    The area of temporally fine-grained video representation learning focuses on generating frame-by-frame representations for temporally dense tasks, such as fine-grained action phase classification and frame retrieval. In this work, we advance the state-of-the-art for self-supervised models in this area by re-examining the design of transformer architectures for video representation learning. A key aspect of our approach is the improved sharing of scene information in the temporal pipeline by representing multiple salient entities per frame. Prior works use late-fusion architectures that reduce frames to a single-dimensional vector before modeling any cross-frame dynamics. In contrast, our \textbf{M}ulti-entity \textbf{V}ideo Trans\textbf{former} (\textbf{MV-Former}) processes the frames as groups of entities represented as tokens linked across time. To achieve this, we propose a Learnable Spatial Token Pooling strategy to identify and extract features for multiple salient regions per frame. Through our experiments, we show that MV-Former outperforms previous self-supervised methods, and also surpasses some prior works that use additional supervision or training data. When combined with additional pre-training data from Kinetics-400, MV-Former achieves a further performance boost. Overall, our MV-Former achieves state-of-the-art results on multiple fine-grained video benchmarks and shows that parsing video scenes as collections of entities can enhance performance in video tasks.
\end{abstract}
\section{Introduction}

\begin{figure}[t]
  \centering
   \includegraphics[width=1.0\linewidth]{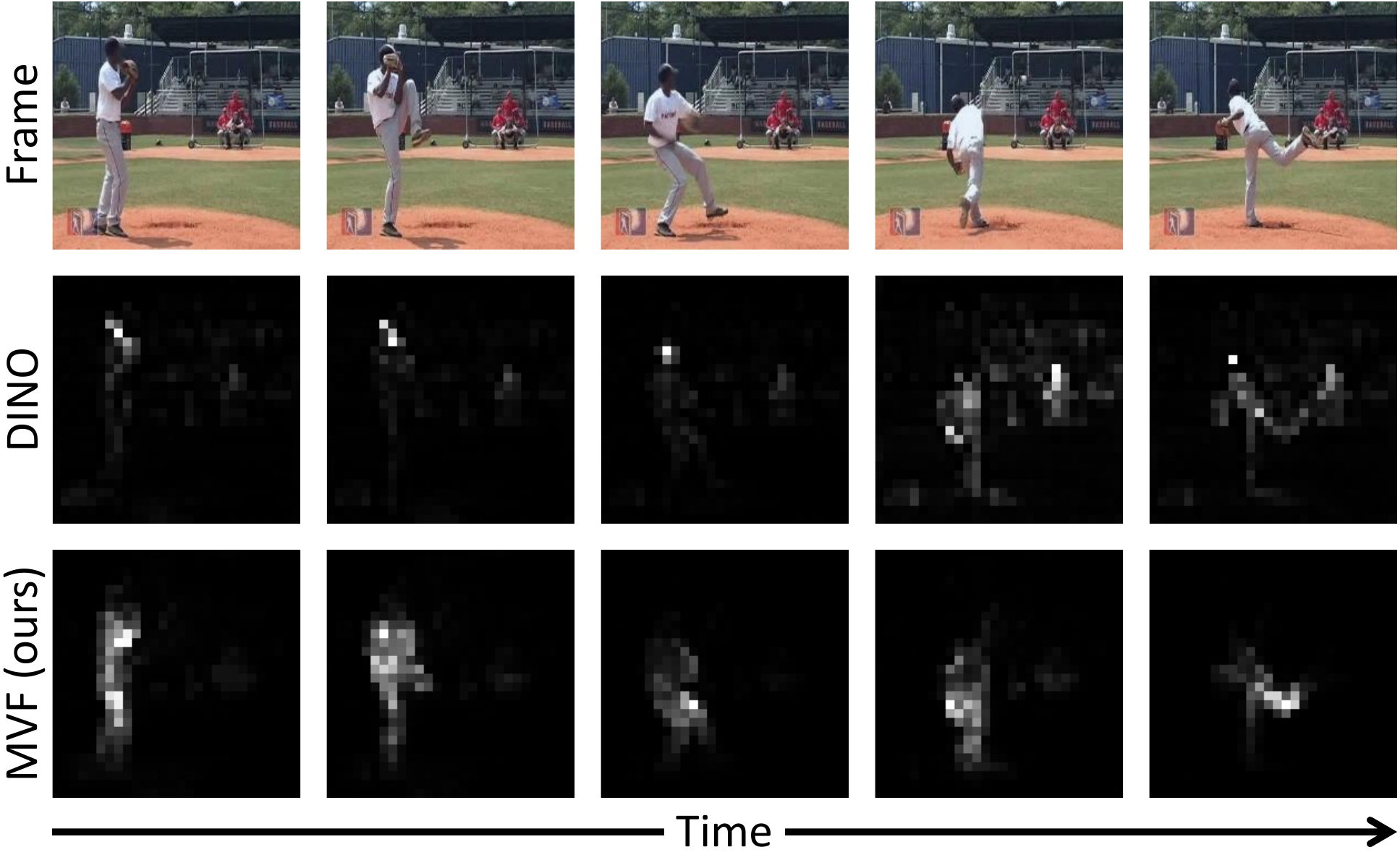}
   \vspace{-0.2in}
   \caption{We present \textbf{MV-Former}, a \textbf{M}ulti-entity \textbf{V}ideo Trans\textbf{former} architecture for fine-grained video representation learning. Our multi-entity approach naturally learns to isolate the main actor in a video using only contrastive self-supervision. As shown above, MV-Former attention has superior actor segmentation over DINO. MV-Former also achieves state of the art performance on multiple fine-grained video datasets and tasks.}
   \label{fig:teaser}
   \vspace{-0.1in}
\end{figure}

Self-Supervised Learning (SSL) has been a rapidly growing area of research, showing potential both to scale to massive data and to learn in environments where annotations are limited or expensive to generate \cite{bachman2019learning, caron2021emerging, chen2020simple, chen2020improved, he2022masked, he2020momentum, hjelm2018learning, misra2020self, zbontar2021barlow, balestriero2023cookbook}. Video Representation Learning is a promising match for SSL \cite{schiappa2023self, selva2023video}, though the majority of existing research in this space focuses on learning a single, video-level representation \cite{feichtenhofer2021large, qian2021spatiotemporal, feichtenhofer2022masked, tong2022videomae, wang2022bevt, tan2021vimpac, recasens2021broaden, wang2022long, dave2022tclr}. However, there are many fine-grained video tasks which require not only video-level understanding, but also a temporally dense understanding too. Such tasks include action phase classification \cite{dwibedi2019temporal, kuehne2014language, sigurdsson2016hollywood, zhang2013actemes}, fine-grained frame retrieval \cite{dwibedi2019temporal, haresh2021learning}, and video temporal alignment \cite{dwibedi2019temporal, cao2020few, hadji2021representation, haresh2021learning}. Methods designed for such tasks focus on generating frame-wise features that are expressive and discriminative not only for the high-level actions in a video, but also the subtle moment-to-moment steps of those actions \cite{misra2016shuffle, sermanet2018time, dwibedi2019temporal, haresh2021learning, hadji2021representation, chen2022frame, zhang2023modeling}. SSL is especially desirable for this domain, as frame-level video annotations are rare and expensive to create.

\begin{figure*}[t!]
   \centering
   \includegraphics[width=1.0\linewidth]{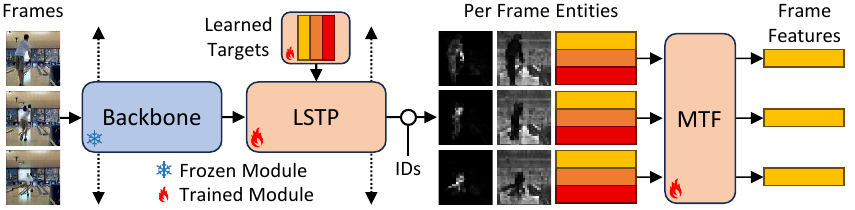}
   \vspace{-0.18in}
   \caption{Overview of our \textbf{MV-Former} architecture. Multi-layer features are extracted from a frozen backbone and fed into our \textbf{L}earnable \textbf{S}patial \textbf{T}oken \textbf{P}ooling (\textbf{LSTP}) module which uses learned embedding targets to extract multiple entities per frame. Then our \textbf{M}ulti-entity \textbf{T}emporal \textbf{F}usion (\textbf{MTF}) transformer combines these entities to produce the final per-frame features. These temporally dense features are then applied to fine-grained downstream tasks including action phase classification and fine-grained frame retrieval.}
   \label{fig:arch}
\vspace{-0.1in}
\end{figure*}

Networks for fine-grained video learning typically use a two-stage approach. First, a frame-level encoder backbone is applied to reduce each frame into a single vector. Second, a temporal fusion module is applied, which allows information to flow between the representations of separate frames. Early works in this field utilized 3D convolutional networks to perform temporal fusion \cite{carreira2017quo, tran2015learning, feichtenhofer2019slowfast, taylor2010convolutional}, but recent works have shifted to transformer-based temporal fusion \cite{chen2022frame, zhang2023modeling}. Transformer fusion allows models to learn long-range inter-frame dynamics through self-attention. However, past architectures are still limited in how they share information over the temporal axis. Prior approaches reduce each frame to a single 1D vector before any inter-frame information is combined. We hypothesize that this bottleneck restricts the ability of models to represent multiple entities, which limits their capacity to learn the temporal dynamics of a scene.

In this work, we re-examine the design of transformer-based architectures for self-supervised video representation learning and propose a new \textbf{M}ulti-entity \textbf{V}ideo Trans\textbf{former} (\textbf{MV-Former}) which achieves state-of-the-art performance on multiple fine-grained video tasks. Central to our approach for MV-Former is the choice to not parse videos as collections of frames, but instead as collections of salient entities, such as the primary actor and the scene background. We call this method \textbf{Multi-entity Temporal Fusion} (\textbf{MTF}). Our approach is based on the intuition that videos contain multiple elements with distinct temporal dynamics. For example, the main human actor in an action recognition video may move rapidly, while the background moves very little. We extract multiple entities per frame with consistent semantics using a \textbf{Learnable Spatial Token Pooling} (\textbf{LSTP}) strategy. MV-Former is able to effectively identify the primary actor in a scene using only contrastive self-supervision. As shown in Figure \ref{fig:teaser}, MV-Former's entity attention maps better segment the main actor than the best DINO CLS attention head. Overall, our MV-Former architecture advances the state-of-the-art for fine-grained tasks on the Penn Action \cite{sermanet2018time}, Pouring \cite{sermanet2018time}, and FineGym \cite{shao2020finegym} datasets, and we demonstrate further improvements when combined with large-scale pre-training on Kinetics-400 \cite{carreira2017quo}. In summary, our contributions are as follows:

\begin{itemize}
    \item \textbf{MV-Former}, a Multi-entity Video Transformer architecture designed for self-supervised fine-grained video representation learning.
    \item \textbf{Multi-entity Temporal Fusion}, a video representation learning approach which fuses information across time by first parsing a scene into a collection of salient entities.
    \item A \textbf{Learnable Spatial Token Pooling strategy to identify and extract multiple salient entities per frame with consistent semantics across time from learned targets.}
    \item Experiments with MV-Former achieving state-of-the-art results on several fine-grained video benchmarks.
\end{itemize}
\section{Related Work}

\myparagraph{Self-Supervised Visual Learning.}
Significant advances have been made in recent years in learning self-supervised visual representations, with many works focused on the recently popularized Vision Transformer (ViT) architecture \cite{dosovitskiy2020image}. As of writing, the most popular methods are contrastive or joint-embedding-based methods \cite{caron2021emerging, oquab2023dinov2, he2020momentum, chen2020simple, grill2020bootstrap}, and masked-reconstruction methods \cite{he2022masked, bao2021beit}.
Many works have been proposed that translate advances in image-level Self-Supervised Learning to the video domain.
\cite{feichtenhofer2021large} demonstrated a unified framework to extend methods like MoCo \cite{he2020momentum}, BYOL \cite{grill2020bootstrap}, SimCLR \cite{chen2020simple}, and SwAV \cite{caron2020unsupervised} into video-level methods by learning to maximize the similarity of representations of different clips from the same video. 
Several works focus on applying masked-reconstruction-based methods to video through strategies like temporal masking and reconstruction \cite{feichtenhofer2022masked, tong2022videomae, wang2023videomae, wang2022bevt, tan2021vimpac}.
Other works focus on learning temporally stable representations by combining wide and narrow views of time \cite{recasens2021broaden, wang2022long, feichtenhofer2019slowfast, dave2022tclr, kuang2021video, ranasinghe2022self, xiao2022hierarchical, qing2022learning}.
Note that these methods usually focus on learning video-level representations that are invariant to changes over time. While such representations are beneficial for high-level understanding, they are not helpful for tasks that require fine-grained, frame-level understanding.

\myparagraph{Fine-Grained Video Representation Learning.}
Works in this domain of research aim to generate dense, frame-level features to enabled video tasks such as fine-grained frame retrieval \cite{dwibedi2019temporal, haresh2021learning}, per-frame action classification \cite{dwibedi2019temporal, kuehne2014language, sigurdsson2016hollywood, zhang2013actemes}, and video alignment \cite{dwibedi2019temporal, cao2020few, hadji2021representation, haresh2021learning}. Much of the research in this field focuses on either self-supervised or weakly-supervised learning methods to reduce or remove the need for costly temporally dense annotations. \cite{sermanet2018time} proposed Time-Contrastive Networks (TCN) and established the Pouring dataset, while \cite{dwibedi2019temporal} created Temporal Cycle-Consistency (TCC) learning and also created additional annotations and benchmarking procedures for the Pouring and Penn Action \cite{zhang2013actemes} datasets. \cite{chen2022frame} improved self-supervised performance by proposing Sequence Contrastive Loss (SCL) and also by applying transformer-based temporal fusion. \cite{zhang2023modeling} further proposes a statistically-motivated learning objective using Brownian Bridges. \cite{chen2022frame} and \cite{zhang2023modeling} use similar video learning architectures, which use a pre-trained frame-level encoder which greatly compresses frame entity information before sharing it with a video-level temporal fusion transformer. In this work, we revisit the design of video learning architectures, and present a new Multi-entity Temporal Fusion architecture with MV-Former.
\section{MV-Former Architecture}

\subsection{Design Motivation}

We propose a \textbf{M}ulti-entity \textbf{V}ideo Trans\textbf{former} architecture (\textbf{MV-Former}), which is designed to parse video scenes not as individual frames, but instead as collections of entities. In this work, we use the term ``entity'' to describe any region of a video frame with a shared and consistent semantic meaning for purposes of scene parsing. Under this definition, an entity can describe a person, an object, or even the entire background of the scene. Though the background may be composed of many separate objects (floors, walls, furniture, \textit{etc.}) for purposes of parsing the video it can often be grouped together as a single entity. In our MV-Former architecture, we extract multiple entities per frame and associate them across time through a shared ID-vector. Our motivation for this approach is based on the intuition that video scenes contain multiple entities with distinct temporal dynamics. For example, in a video of a human performing an action, the person's pose is often highly dynamic, while the scene background is largely static. MV-Former can be separated into three main components: the per-frame Visual Backbone, the Learnable Spatial Token Pooling module, and the Multi-entity Temporal Fusion module. We show these components in Figure \ref{fig:arch} and discuss them in more detail in the following sections.

\subsection{Backbone and Frame-Level Features}

Like prior works, we start by applying a pre-trained backbone on a frame-by-frame basis. However, unlike prior works, our goal for this backbone is not to extract a single compressed frame feature but instead to extract features for multiple salient entities. For this reason, we select a DINO (v1) ViT backbone \cite{caron2021emerging}, and specifically use DINO ViT-B/8, as its larger size and smaller patch resolution provide finer detail for local object features. ViTs produce powerful representations both at a global level and at a local level in the form of spatial token features. Prior works have demonstrated that these local features align well with object boundaries and semantics \cite{caron2021emerging}, so they are well-suited for extracting features for multiple entities per frame. It has also been shown that the alignment of DINO feature semantics with local objects and object parts varies depending on their depth in the network \cite{walmer2023teaching}. For this reason, we use a multi-layer feature extraction strategy and take spatial token features from multiple intermediate layers. In addition, in Section \ref{sec:ablations} we present an ablation demonstrating that LSTP and MTF are also beneficial for a ResNet \cite{he2016deep} backbone.

\subsection{Learnable Spatial Token Pooling}

After the frame-level backbone, MV-Former must identify and extract multiple entities from the spatial token features. Cross-attention is a desirable mechanism for this purpose, as it can flexibly and dynamically extract features from regions of different shapes and sizes. We draw inspiration from \cite{mukhoti2023open} which uses the text encoder of a CLIP model to guide self-supervised segmentation through cross-attention on the visual token features. As we are working with DINO, a vision-only model, there are no language encoder features to guide this cross-attention. Instead, we propose a \textbf{Learnable Spatial Token Pooling (LSTP)}, which uses learnable embedding vectors for the cross-attention input. These embedding vectors are trained as parameters along with the rest of the network, and they allow the network to learn which features are worth extracting from the scene. Our approach has some similarity with Q-Former \cite{li2023blip}, as both use cross-attention with learnable queries to extract information from a ViT backbone. However, while Q-Former is designed to bridge vision and language representations, our LSTP instead is created to extract salient per-frame visual entities for temporal fusion in video. The number of learnable embedding vectors determines the number of entities extracted per frame. In our primary results, we use 3 or 6 entities per frame depending on the dataset. We also present an ablation in Section \ref{sec:ablations} comparing different entity counts. These learnable embedding targets are held constant along the temporal dimension, which enables the LSTP module to extract consistent features for the most salient objects and image regions. In practice, LSTP naturally learns to segment out the primary actor in the scene and the scene background, as we show in Section \ref{sec:vis}.

\begin{table*}[t!]
 \setlength{\cmidrulewidth}{0.01em}
\renewcommand{\tabcolsep}{4pt}
\renewcommand{\arraystretch}{1.1}
\caption{Experimental results for three fine-grained video datasets. We achieve state-of-the-art self-supervised results on all metrics for Penn Action and FineGym, and in fine-grained frame retrieval for Pouring. For details on tasks and metrics, please see Section \ref{sec:expmeth}.}
\vspace{-0.8em}
\centering
\footnotesize
\resizebox{\linewidth}{!}{
\begin{tabular}{@{}lllllllllll@{}}
\toprule
 &  \multicolumn{4}{c}{\textbf{Penn Action}} & \multicolumn{4}{c}{\textbf{Pouring}} &  \multicolumn{2}{c}{\textbf{FineGym}} \\
 \cmidrule[\cmidrulewidth](l){2-5} \cmidrule[\cmidrulewidth](l){6-9} \cmidrule[\cmidrulewidth](l){10-11}
Method & Class. & Progress & Tau & Retrieval & Class. & Progress & Tau & Retrieval & Class. (99) & Class. (288)\\
 \midrule
 CARL \cite{chen2022frame}     & 93.07 & 0.918 & 0.985 & 92.28     & 93.73 & 0.935 & \textbf{0.992} & -     & 41.75 & 35.23             \\
 VSP \cite{zhang2023modeling}  & 93.12 & 0.923 & 0.986 & 92.56     & \textbf{93.85} & \textbf{0.942}  & 0.990 & 91.85   & 43.12 & 36.95             \\
 \midrule
 MVF (ours) & \textbf{94.21}\stdev{0.04} & \textbf{0.931}\stdev{0.006} & \textbf{0.989}\stdev{0.002} & \textbf{92.99}\stdev{0.06} &       93.34\stdev{2.00} & 0.919\stdev{0.015} & 0.985\stdev{0.003} & \textbf{92.33}\stdev{1.02}             & \textbf{44.77}\stdev{0.71} & \textbf{38.30}\stdev{0.26}\\
\bottomrule
\end{tabular}
  }
\label{tab:penn_fine}
\end{table*}

\subsection{Multi-Entity Temporal Fusion}
\label{sec:archmtf}

Finally, we fuse the per-frame per-entity features across the temporal dimension using a \textbf{Multi-entity Temporal Fusion (MTF)} module. The purpose of this module is to generate dense, per-frame features that are enriched through temporal context. Like \cite{chen2022frame}, we use a three block transformer to perform fusion of the frame-level features. However, rather than feed in a single token per frame, we input multiple tokens per frame to represent the multiple entities extracted through LSTP. To differentiate the entities, we append a one-hot ID vector to the end of each entity feature vector during token generation. For a standard transformer architecture, the number of input tokens will be equal to the number of output tokens, meaning a separate feature is generated per input entity. To reduce this to a single output feature per frame, we consider two options. The first is to simply take the average of all the per-frame tokens. The second is a ``CLS-token-style'' approach, where we designate one token of each frame to act as the output token. This approach provides greater flexibility than average pooling, which tends to provide similar gradients to each of the separate entity tokens. We find that the final pooling choice may favor some metrics over others. Specifically, we find that the averaging approach performs better for the Classification and Retrieval metrics, while the CLS-style approach works better for Phase Progression. Also note that under our multi-entity approach, the effective ``width'' of the transformer is multiplied by the number of entities, however, this does not increase the number of parameters. To provide a uniform baseline of comparison, we also present a ``fixed-width'' baseline in Section \ref{sec:ablations} which simulates the increased width of our MTF module.
\section{Experimental Methods}
\label{sec:expmeth}

\myparagraph{Datasets.}
We follow the protocols of \cite{dwibedi2019temporal} and \cite{chen2022frame} and conduct benchmarking experiments on three video datasets: Penn Action \cite{zhang2013actemes}, Pouring \cite{sermanet2018time}, and FineGym \cite{shao2020finegym}. For FineGym, we conduct experiments on two different splits, FineGym99 and FineGym288. FineGym288 has additional category labels and training data. All of our experiments are trained self-supervised without labels, so for FineGym288 we utilize the extra training data but not the extra label information. Additionally, we conduct further experiments with large-scale pre-training on Kinetics-400 \cite{carreira2017quo}.

\myparagraph{Tasks and Metrics.}
We assess MV-Former using four standard tasks and metrics. 
\textbf{(1) Fine-Grained Phase Classification: } In this task, videos are annotated on a frame-by-frame level by dividing the actions into key phases. After self-supervised pre-training, the model is frozen and a linear classifier is trained to predict the action phases. Classification accuracy is reported.
\textbf{(2) Phase Progression: } For each frame, the model must predict how much time is left until the next action phase boundary. A linear regression model is trained on top of the frozen network, and the average R-squared metric is reported.
\textbf{(3) Kendall's Tau: } Given two pairs of frames from two different videos, the model must match the frames such that the pairs have the same temporal ordering. This is achieved through nearest neighbors matching. Fraction of correct matches is reported.
\textbf{(4) Fine-Grained Frame Retrieval: } Given a query frame, return k frames with the same fine-grained frame action label as the query. We report results for Average Precision with $k=5$ (AP@5). For additional details, please see \cite{dwibedi2019temporal, chen2022frame}.

\myparagraph{Baselines.}
We compare MV-Former with the most recent state-of-the-art methods for fine-grained self-supervised video representation learning, which are CARL \cite{chen2022frame} and VSP \cite{zhang2023modeling}. Over the years, many methods have been proposed that use different levels of weakly supervised data, including paired video samples \cite{dwibedi2019temporal, haresh2021learning, liu2022learning, bar2024weakly}, or phase boundary time stamps \cite{zhang2023modeling}. We present a comprehensive comparison with additional self-, weakly-, and fully-supervised methods in Section \ref{sec:addsup}. We also note that some works focus on learning video representations for human action recognition on pre-extracted sequences of human skeleton joint positions in place of or in addition to RGB frames \cite{duan2022revisiting, yan2023skeletonmae}. Such works show promising results, but they are also reliant on joint extraction models that are trained with extensive annotations. In this work, we focus on comparing with video representation learning methods that operate in RGB space.

\myparagraph{Training Details.}
For our frame-level backbone, we utilize DINO ViT-B/8 \cite{caron2021emerging}. We find that some datasets benefit from features derived from earlier layers, while others prefer features derived from only later layers. We extract features from layers 4, 8, and 12 for Penn Action and Kinetics-400, and from layers 10, 11, and 12 for FineGym. For Pouring, we use only the final layer features, as using extra features is detrimental due to the very small size of the dataset. For Penn Action, Pouring, and Kinetics-400, we use 3 entities per frame, and for FineGym we increase this to 6 entities per frame. We use a CLS-style final feature selection strategy for Penn Action, Pouring, and Kinetics-400, and we use the average-pooling strategy for FineGym classification on both splits. We train MV-Former using Sequence Contrastive Loss (SCL) \cite{chen2022frame}. For Pouring, Penn Action, FineGym99/288, and Kinetics-400, we train for 1000, 500, 300, and 10 epochs respectively. On Penn Action and Pouring, we train with batch size 4 on 4 A100 GPUs, and for FineGym and Kinetics-400 pre-training we use batch size 8 on 8 A100 GPUs.

\begin{table}[!t]
\setlength{\cmidrulewidth}{0.01em}
\renewcommand{\tabcolsep}{4pt}
\renewcommand{\arraystretch}{1.1}
\caption{Results for our MV-Former and baselines on Penn Action with Kinetics-400 pre-training. MV-Former achieves state-of-the-art performance in Classification and Retrieval. The highest score for each metric is \textbf{bold} and the second highest is \underline{underlined}.}
\vspace{-0.8em}
\centering
\footnotesize
\resizebox{1.01\linewidth}{!}{
\begin{tabular}{@{}lccllll@{}}
    \toprule
    Method & Pre. & Fine. & Class. & Progress & Tau & Retrieval\\
    \midrule
    CARL & \multicolumn{2}{c}{w/o Pre.} & 93.07 & 0.918 & 0.985 & 92.28\\
    CARL & \checkmark & & 91.9 & 0.903 & 0.949 & -\\
    CARL & \checkmark & \checkmark & {93.9} & 0.908 & 0.977 & -\\
    VSP & \multicolumn{2}{c}{w/o Pre.} & 93.12 & 0.923 & 0.986 & 92.56\\
    VSP & \checkmark & & 92.35 & 0.894 & 0.952 & -\\
    VSP & \checkmark & \checkmark & 93.57 & \textbf{0.944} & \underline{0.988} & -\\
    \midrule
    MVF & \multicolumn{2}{c}{w/o Pre.} & \underline{94.21}\stdev{0.04} & \underline{0.931}\stdev{0.006} & \textbf{0.989}\stdev{0.002} & \underline{92.99}\stdev{0.06}\\
    MVF & \checkmark & & 91.62\stdev{0.30} & 0.927\stdev{0.005} & 0.895\stdev{0.004} & 88.38\stdev{0.20}\\
    MVF & \checkmark & \checkmark & \textbf{94.56}\stdev{0.32} & 0.924\stdev{0.004} & 0.980\stdev{0.002} & \textbf{93.40}\stdev{0.08}\\
    \bottomrule
\end{tabular}
\label{tab:k400}
  }
\vspace{-0.1in}
\end{table}

\myparagraph{Codebase.} Our codebase is built on the Contrastive Action Representation Learning framework \cite{chen2022frame}, and can be found at: \url{https://github.com/facebookresearch/video_rep_learning}. We have maintained exact experiment configuration files tracking all hyperparameters to enable full reproducibility of all results presented.

\myparagraph{Measuring the Impact of Initialization.}
For our evaluation protocols, we make one major change from prior works, as we choose to measure and report results for multiple random initializations per model configuration and dataset. For any optimization procedure, the final state of the network will depend on the initial state, but a well-performing network and objective together should converge to a good solution consistently, regardless of the initial state. We believe it is important to consider the impact of the random network initialization on the quality of the final network. While prior works report results for only one trial per model, we instead adopt a multi-trial protocol. Specifically, in each test we conduct three trials with different random initialization seeds, and we report all results as the mean plus/minus two standard deviations. Through this protocol, we show that MV-Former achieves state-of-the-art performance on Penn Action and FineGym by a statistically significant margin. This also allows us to measure the variance of the four commonly used benchmark tasks and metrics. We find that the Phase Progression task and metric has the highest sensitivity to the model initialization. We encourage future works to adopt a similar multi-trial methodology.

\begin{table}[!t]
\setlength{\cmidrulewidth}{0.01em}
\renewcommand{\tabcolsep}{6.38pt}
\renewcommand{\arraystretch}{1.145}
\caption{Ablation study of MV-Former design details on Penn Action. ``FWB'' is a Fixed-Width Baseline which emulates the extra width of MTF without LSTP. ``Last'' uses only last-layer features instead of multi-layer features. ``Ent.'' denotes the entity count.}
\vspace{-0.8em}
\centering
\footnotesize
\resizebox{1.01\linewidth}{!}{
\begin{tabular}{@{}llllll@{}}
\toprule
Feats. & Ent. & Class. & Progress & Tau & Retrieval\\
\midrule
FWB & 3* & 88.28\stdev{0.48} & 0.913\stdev{0.008} & 0.995\stdev{0.0002} & 86.49\stdev{0.27}\\
FWB & 5* & 87.97\stdev{0.12} & 0.911\stdev{0.004} & 0.995\stdev{0.0001} & 86.26\stdev{0.12}\\
\midrule
Last & 1 & 93.35\stdev{0.61} & 0.924\stdev{0.006} & {0.990}\stdev{0.0001} & 91.97\stdev{0.79}\\
Last & 3 & 93.92\stdev{0.24} & {0.926}\stdev{0.004} & 0.989\stdev{0.0003} & 92.81\stdev{0.30}\\
Last & 5 & 93.50\stdev{0.27} & 0.911\stdev{0.006} & \textbf{0.992}\stdev{0.001} & 92.50\stdev{0.3}\\
\midrule
Multi. & 1 & \textbf{94.23}\stdev{0.08} & 0.920\stdev{0.007} & 0.987\stdev{0.0005} & {93.02}\stdev{0.18}\\
Multi. & 3 & {94.21}\stdev{0.04} & \textbf{0.931}\stdev{0.006} & 0.989\stdev{0.002} & 92.99\stdev{0.06}\\
Multi. & 5 & 94.11\stdev{0.18} & {0.926}\stdev{0.005} & 0.989\stdev{0.002} & \textbf{93.07}\stdev{0.01}\\
\bottomrule
\end{tabular}
\label{tab:ablateef}
  }
\vspace{-0.1in}
\end{table}
\section{Results}
\label{sec:results}

\begin{table*}[!t]
 \setlength{\cmidrulewidth}{0.01em}
\renewcommand{\tabcolsep}{10.5pt}
\renewcommand{\arraystretch}{0.95}
\caption{Comprehensive comparison with methods that use self-, weakly-, or fully-supervised learning.
Our MV-Former is only surpassed by other methods that use additional supervision or data.
The highest score for each metric is \textbf{bold} and the second highest is \underline{underlined}.
}
\vspace{-0.8em}
\centering
\footnotesize
\resizebox{1.0\linewidth}{!}{
    \begin{tabular}{@{}lllllllll@{}}
    \toprule
    & & & \multicolumn{4}{c}{\textbf{Penn Action}} & \multicolumn{2}{c}{\textbf{FineGym}} \\
    \cmidrule[\cmidrulewidth](l){4-7} \cmidrule[\cmidrulewidth](l){8-9}
    Method & Labels & Pretrain & Class. & Progress & Tau & Retrieval & Class. (99) & Class. (288) \\
    \midrule
    TCC* \cite{dwibedi2019temporal} & Video & - & 81.35 & 0.664 & 0.701 & 76.74 & 25.18 & 20.82 \\
    TCC \cite{dwibedi2019temporal} & Video & - & 74.39 & 0.591 & 0.641 & - & - & - \\
    TCC$^\dagger$ \cite{dwibedi2019temporal, chen2022frame} & Video & - & 86.35 & 0.899 & 0.980 & - & - & - \\
    LAV* \cite{haresh2021learning} & Video & - & 84.25 & 0.661 & 0.805 & 79.13 & - & - \\
    LAV \cite{haresh2021learning} & Video & - & 78.68 & 0.625 & 0.684 & - & - & - \\
    GTA \cite{hadji2021representation} & Video & - & - & 0.789 &  0.748 & - & - & - \\
    VAVA \cite{liu2022learning} & Video & - & 84.48 & 0.709 & 0.805 & - & - & - \\
    LRProp \cite{bar2024weakly} & Video & - & 93.25 & 0.930 &  \textbf{0.991} & 92.46 & - & - \\
    \midrule
    SaL \cite{misra2016shuffle} & None & - & 68.15 & 0.390 & 0.474 & 76.04 & 21.45 & 19.58 \\
    TCN \cite{sermanet2018time} & None & - & 68.09 & 0.383 & 0.542 & 77.84 & 20.02 & 17.11 \\
    TCN$^\dagger$ \cite{sermanet2018time, chen2022frame} & None & - & 86.31 & 0.898 & 0.832 & - & - & - \\
    CARL \cite{chen2022frame} & None & - & 93.07 & 0.918 & 0.985 & 92.28 & 41.75 & 35.23 \\
    VSP \cite{zhang2023modeling} & None & - & 93.12 & 0.923 & 0.986 & 92.56 & 43.12 & 36.95 \\
    \midrule
    VSP-P \cite{zhang2023modeling} & Phase & - & 93.27 & - & - & \underline{93.45} & 44.58 & 38.23 \\
    VSP-F \cite{zhang2023modeling} & Frame & - & \underline{94.24} & - & - &\textbf{94.89} & \textbf{45.66} & \textbf{39.48} \\
    \midrule
    CARL \cite{chen2022frame} & None & K-400 & 93.9 & 0.908 & 0.977 & - & - & - \\
    VSP \cite{zhang2023modeling} & None & K-400 & 93.57 & \textbf{0.944} & 0.988 & - & - & - \\
    \midrule
    MVF (ours) & None & - & {94.21}\stdev{0.04} & \underline{0.931}\stdev{0.006} & \underline{0.989}\stdev{0.002} & {92.99}\stdev{0.06} & \underline{44.77}\stdev{0.71} & \underline{38.30}\stdev{0.26} \\
    MVF (ours) & None & K-400 & \textbf{94.56}\stdev{0.32} & 0.924\stdev{0.004} & 0.980\stdev{0.002} & {93.40}\stdev{0.08} & - & - \\
    \bottomrule
    \end{tabular}
}
\label{tab:all_penn_fine}
\end{table*}

\subsection{Fine-Grained Datasets}
\label{sec:bench}

We summarize our results for MV-Former on three fine-grained video representation learning benchmarks, Penn Action, Pouring, and FineGym, with results in Table \ref{tab:penn_fine}.

\textbf{Penn Action.}
MV-Former achieves state-of-the-art self-supervised performance in all four metrics for Penn Action. The largest gain is achieved in Classification, with an increase of $1.09\%$ over VSP. For all four metrics, our improvement is at least two standard deviations above the best performing prior work. We note that Phase Progression has the highest standard deviation relative to its absolute value, showing that it is the most sensitive to the random network initialization. Meanwhile, Classification and Retrieval both have relatively low sensitivity to initialization.

\textbf{Pouring.}
For Pouring we do not see consistent improvement for all metrics, and in addition we see much higher variance across trials for all four metrics as compared to Penn Action. We attribute both of these issues to the small size of the Pouring training set, which contains only 70 videos, making it roughly 16 times smaller than Penn Action and 45 times smaller than FineGym99. However, MV-Former does achieve state-of-the-art results for Fine-Grained Frame Retrieval on this benchmark.

\textbf{FineGym.}
Like prior works, we report fine-grained classification accuracy on two splits, FineGym99 and FineGym288. MV-Former achieves state-of-the-art self-supervised results for Phase Classification on both splits. MV-Former's average score surpasses VSP by $1.65\%$ for FineGym99 and $1.35\%$ for FineGym288. These results surpass the prior bests by at least two standard deviations.

\subsection{Large-Scale Pre-training}

We show that MV-Former can achieve further performance improvements on the Penn Action dataset through large-scale pre-training on Kinetics-400. We again follow our 3-trial experimental protocol and pre-train three different MV-Formers on Kinetics-400 for 10 epochs, and then fine-tune them on Penn Action for 500 epochs. The results are summarized in Table \ref{tab:k400}. We see that Kinetics-400 pre-training boosts Classification performance by another $0.35\%$. We also see a $0.41\%$ gain in Retrieval over our base MV-Former.  We find that Kinetics-400 pre-training is not beneficial for Progress and Tau, which matches previous trends observed by \cite{chen2022frame} for CARL.

\subsection{Methods using Additional Supervision}
\label{sec:addsup}

In Table \ref{tab:all_penn_fine} we present a comprehensive, unified comparison of recent approaches on the Penn Action and FineGym datasets for fine-grained video representation tasks. In addition to self-supervised methods, we also compare with methods using weak or full supervision, and for Penn Action we also include methods using Kinetics-400 pre-training. ``Video labels'' means the method requires a video-level label of the action category in order to sample video pairs of the same category. ``Phase labels'' means the method uses annotations for the positions of the action phase boundaries in time, but not the action phase categories. ``Frame labels'' means the method uses fully labelled data with phase boundaries and phase category labels. Methods TCC* and LAV* denote versions where a separate model is trained for each Penn Action category. TCC$^\dagger$ and TCN$^\dagger$ denote reimplemented versions presented by \cite{chen2022frame} which utilize ViT-based temporal fusion architectures. MV-Former with Kinetics-400 pre-training achieves the overall highest performance in Penn Action Phase Classification, surpassing even the fully-supervised version of VSP (VSP-F). For FineGym classification, MV-Former is second only to the fully-supervised VSP-F. Overall, MV-Former is only surpassed by methods that use additional supervision or training data.

\begin{figure*}[t!]
  \centering
   \includegraphics[width=1.0\linewidth]{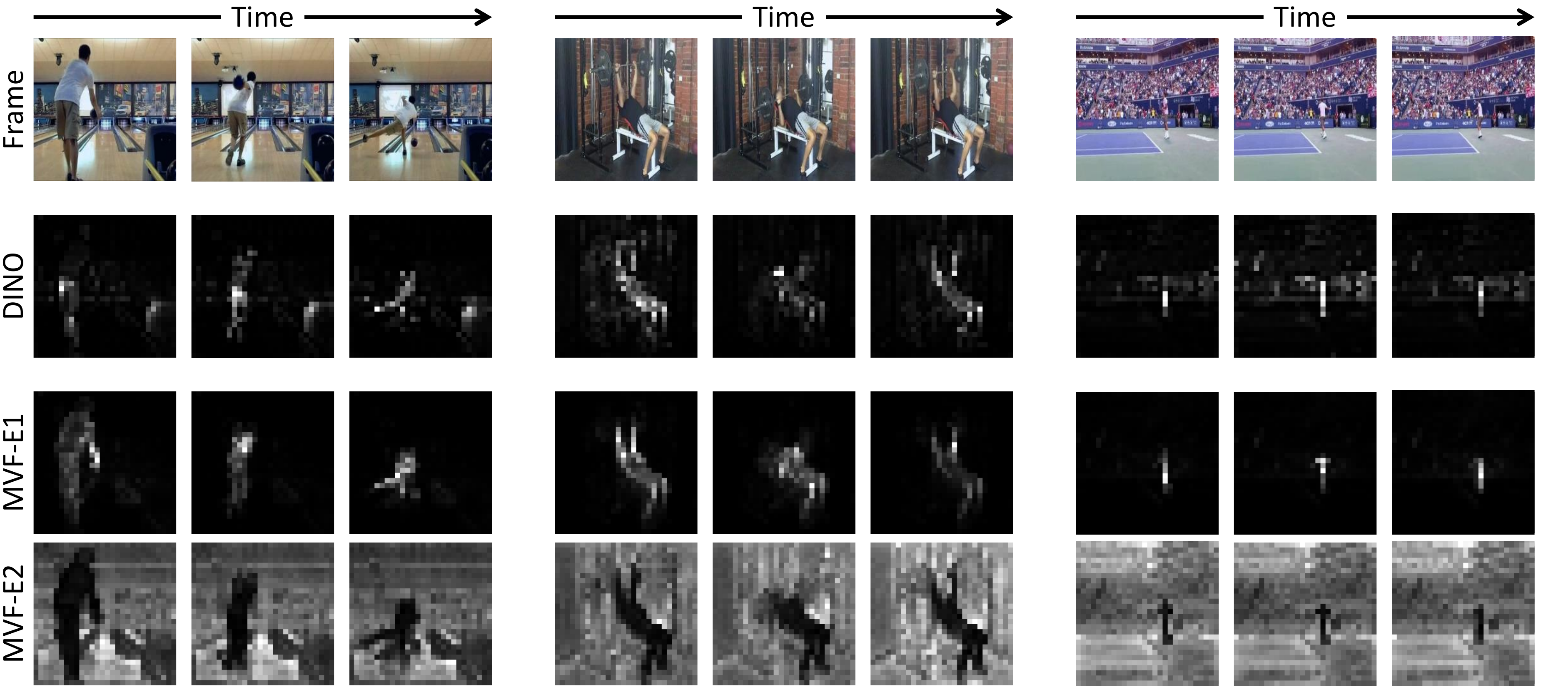}
   \vspace{-0.15in}
   \caption{Visualizations of the attention maps of our LSTP module. Our first entity (MVF-E1) learns to attend to the primary actor, with a particular focus on the limbs. Our second entity (MVF-E2) focuses on the scene background. We compare with the most actor-aligned CLS attention head of DINO ViT-B/8. Our entity-learning approach better isolates and segments the primary actor in the scene over DINO.}
   \label{fig:attvis}
   \vspace{-0.05in}
\end{figure*}

\subsection{Visualizing Learnable Spatial Token Pooling}
\label{sec:vis}

We visualize the attention maps created by our LSTP module, which correspond to the regions that contribute to each entity's features. We take several sample videos from the Penn Action dataset and visualize the LSTP attention maps for our best performing MV-Former model in Figures \ref{fig:teaser} and \ref{fig:attvis}. In addition, we present a comparison of our LSTP attention maps with DINO ViT-B/8's CLS token attention maps. For sake of fair comparison, we manually select the DINO head that is most aligned with the scene actor. We find that our first entity's attention maps (denoted MVF-E1) are very well-aligned with the primary scene actor throughout each video. Meanwhile, the DINO attention maps also include the actor, but they are typically far noisier and include large portions of the background. In the right example of Figure \ref{fig:attvis}, DINO's attention maps are distracted by the large crowd in the background, while MVF-E1 clearly focuses on the athlete. Under our CLS-style output configuration, the first entity is connected to the final output token, so it is intuitive that it learns to focus on the most important part of the scene. Additionally, in Figure \ref{fig:attvis} we also show examples of entity 2's attention maps, which effectively segment out the scene background. Note that the third entity (not shown) also attends to the primary actor in the scene. Our LSTP implementation includes no restrictions to enforce that entities be disjoint, though this could be an area of future research. These visualizations show the strength of LSTP for isolating the primary actor and background, improving over DINO's segmentation capabilities.

\subsection{Ablations}
\label{sec:ablations}

We present ablations for our backbone and feature pooling methodologies on Penn Action. We can ablate MTF while still using LSTP by extracting only one entity per frame, denoted as ``LSTP-1''. When using LSTP with MTF, the number of entities is 3. We also present ablations for multi-layer features, entity count, and a fixed-width baseline. We primarily focus on the Classification, Progress, and Retrieval metrics, as Tau shows only small variations in performance across configurations. Overall, the scores for Tau are quite close to $1.0$, so we believe minor variations in the Kendall's Tau score are not majorly representative of model quality.

\myparagraph{LSTP and MTF with ResNet-50.}
We demonstrate that LSTP and MTF are beneficial for a ResNet-50 backbone, as shown in Table \ref{tab:ablatern}. For these experiments, we keep the ResNet backbone frozen. In row 1, we present results for a basic ResNet-50 backbone with the same temporal-fusion module as \cite{chen2022frame}. In row 2, the use of LSTP-1 with one entity improves performance in Classification, Progress, and Retrieval, and leaves Tau roughly the same. In row 3, utilizing LSTP with three entities and MTF improves Progress, but slightly degrades Classification and Retrieval. Despite this, ResNet with LSTP and MTF still outperforms the baseline ResNet architecture in all metrics except Tau.

\myparagraph{DINO Feature Extraction Methods.} We test how our complete pipeline compares with a simpler model that uses one DINO feature token per frame with the same fusion transformer as \cite{chen2022frame}. We test three traditional methods for DINO feature selection: max pooling, average pooling, and the CLS token. For fair comparison, our methods use only final layer features for this analysis. The results are summarized in Table \ref{tab:ablatedino}.
Among the traditional methods, max pooling gives the lowest performance in most metrics, while the CLS token tends to be the best.
We again compare with LSTP-1, which is functionally similar to the CLS token, though with the disadvantage of needing to be trained from scratch. LSTP-1 performs similarly to CLS, though it is slightly better in Progress and slightly worse in Classification and Retrieval.
Finally, we compare with our full method, using LSTP and MTF with three entities per frame. Our proposed method provides a clear performance boost over DINO features alone, with the highest scores in Classfication, Progress, and Retrieval.

\begin{table}[!t]
\setlength{\cmidrulewidth}{0.01em}
\renewcommand{\tabcolsep}{4pt}
\renewcommand{\arraystretch}{1.595}
\caption{Ablation study applying LSTP and MTF to a ResNet-50 backbone instead of a DINO ViT backbone. Results presented on Penn Action. ``LSTP-1'' uses LSTP with only 1 learned entity, and thus does not use Multi-entity Temporal Fusion. ``LSTP+MTF'' uses 3 learned entities in this experiment. We find that our proposed methods are also beneficial for a ResNet backbone.}
\vspace{-0.8em}
\centering
\footnotesize
\resizebox{1.01\linewidth}{!}{
\begin{tabular}{@{}lllll@{}}
\toprule
Method & Class. & Progress & Tau & Retrieval\\
\midrule
RN50 & 91.96\stdev{0.48} & 0.900\stdev{0.008} & {0.993}\stdev{0.001} & 90.88\stdev{0.38}\\
RN50 + LSTP-1 & \textbf{92.77}\stdev{0.08} & {0.909}\stdev{0.008} & 0.991\stdev{0.001} & \textbf{91.50}\stdev{0.23}\\
RN50 + LSTP + MTF & {92.46}\stdev{0.24} & \textbf{0.915}\stdev{0.002} & 0.992\stdev{0.0004} & {91.15}\stdev{0.20}\\
\bottomrule
\end{tabular}
\label{tab:ablatern}
}
\vspace{-0.07in}
\end{table}

\myparagraph{Fixed Width Baseline.} We propose a Fixed-Width Baseline (FWB) in order to demonstrate that the improved performance of MTF is not simply a consequence of its increased width.
By processing 3 entities per frame, we effectively make the transformer fusion module 3 times wider in terms of the number of tokens, however, this does not change the number of learnable parameters at all. To create a fixed-width baseline, we simulate the increased width of MTF by splitting the CLS token of DINO into three separate tokens using a linear layer. The new tokens are
fed into the temporal fusion transformer just as they would be for MTF, which gives an architecture of equivalent computational width to our proposed method. We conduct this experiment for 3 or 5 tokens per frame.
The results are shown in Table \ref{tab:ablateef} rows 1 and 2. We find that this simple strategy for increasing the width of the fusion transformer does not improve performance, and actually hinders it. This demonstrates that MTF is not simply benefiting from increased width, and also shows that LSTP is an effective strategy to extract multiple useful entities per frame.

\myparagraph{Entity Count.}
We experiment with our selection of entities extracted per frame for LSTP and MTF. We test configurations with 1, 3, and 5 entities. We present this ablation both with and without the use of multi-layer features, which we discuss more in the following section. These results are shown in Table \ref{tab:ablateef}, with rows 3-5 using last-layer features and rows 6-8 using multi-layer features. To summarize, we find that using 3 entities per frame gives the best all around performance, with a balance of high scores between the four metrics. This is the case both when using multi-layer features and when using last-layer features alone. Increasing the number of entities to 5 or decreasing to 1 can sometimes be slightly beneficial, but is usually detrimental.

\myparagraph{Multi-Layer Features.}
We ablate our use of multi-layer DINO features, as shown by rows 3-8 of Table \ref{tab:ablateef}. We either extract features from the final layer only (Last) or from the ends of blocks 4, 8, and 12 (Multi.). We find that the extra features are generally beneficial for all metrics excluding Tau. This holds for all entity counts tested. While minor differences in the model settings may slightly favor some metrics over others, we find that using 3 entities with multi-layer features is the best all-around performer.

\begin{table}[!t]
\setlength{\cmidrulewidth}{0.01em}
\renewcommand{\tabcolsep}{4pt}
\renewcommand{\arraystretch}{1.0}
\caption{Ablation comparing standard DINO feature aggregation methods (MAX-pool, AVG-pool, CLS token) against our proposed LSTP and MTF approaches on the Penn Action dataset. ``LSTP-1'' uses only one learnable entity and does not use MTF. ``LSTP+MTF'' uses 3 entities for the best all-around performance. For fair comparison, all methods use only last-layer features.}
\vspace{-0.8em}
\centering
\footnotesize
\resizebox{1.01\linewidth}{!}{
\begin{tabular}{@{}lllllll@{}}
\toprule
Method & Class. & Progress & Tau & Retrieval\\
\midrule
DINO + MAX & 87.61\stdev{0.12} & 0.883\stdev{0.005} & \textbf{0.996}\stdev{0.0001} & 83.76\stdev{0.21}\\
DINO + AVG & 92.57\stdev{0.07} & 0.904\stdev{0.012} & {0.995}\stdev{0.001} & 90.52\stdev{0.4}\\
DINO + CLS & {93.48}\stdev{0.16} & 0.918\stdev{0.007} & 0.992\stdev{0.0002} & {92.31}\stdev{0.16}\\
\midrule
DINO + LSTP-1 & 93.35\stdev{0.61} & {0.924}\stdev{0.006} & 0.990\stdev{0.0001} & 91.97\stdev{0.79}\\
DINO + LSTP + MTF & \textbf{93.92}\stdev{0.24} & \textbf{0.926}\stdev{0.004} & 0.989\stdev{0.0003} & \textbf{92.81}\stdev{0.30}\\
\bottomrule
\end{tabular}
\label{tab:ablatedino}
}
\end{table}
\section{Discussion \& Conclusion}
\label{sec:conc}

\myparagraph{Limitations \& Future Work.}
Our MV-Former is effective at isolating the main actor and background. However, it does not consistently isolate salient task-specific objects, like the weights in a weightlifting scene. In the future, we would like to improve this capacity by applying language-based weak supervision. In addition, future work could explore how the MV-Former architecture performs when combined with other video-based self-supervised learning objectives, like TCC, TCN, and VSP. Finally, as discussed in Section \ref{sec:archmtf}, MTF introduces additional tokens in the fusion transformer, which does not increase the parameter count but does increase the compute cost. Future work could potentially address this by performing token merging along the temporal dimension, exploiting the high redundancy of information between frames to reduce the token count.

\myparagraph{Conclusion.} In this work, we have presented MV-Former, a Multi-entity Video Transformer designed for fine-grained video representation learning. MV-Former parses scenes into multiple salient entities before performing transformer-based temporal fusion. We refer to this approach as Multi-entity Temporal Fusion, which is distinct from prior works where each frame must be reduced to a single vector representation before any information is shared between frames. To generate these entities, we have proposed a cross-attention-based Learnable Spatial Token Pooling strategy, which uses trainable query embeddings to learn which information to extract from each frame. We have also shown that this approach naturally learns to separate out the primary actor in each scene and the background with only contrastive self-supervision. MV-Former achieves state-of-the-art results in multiple fine-grained video tasks on the Penn Action, Pouring, and FineGym datasets. We also show how MV-Former benefits from large-scale pre-training on Kinetics-400, further advancing performance in classification and retrieval. Overall, our result with MV-Former show that parsing videos as collections of entities can improve fine-grained video representation learning.

\myparagraph{Acknowledgments.} We would like to thank our peers Chao-Yuan Wu, Sumedha Singla, and Florian Metze for their valuable feedback and suggestions for this work.

{
    \small
    \bibliographystyle{ieeenat_fullname}
    \bibliography{main}
}

\end{document}